# PointSSIM: A novel low dimensional resolution invariant image-to-image comparison metric

Oscar Ovanger, Ragnar Hauge, Jacob Skauvold, Michael J. Pyrcz, Jo Eidsvik

This paper presents PointSSIM, a novel low-dimensional image-to-image comparison metric that is resolution invariant. Drawing inspiration from the structural similarity index measure and mathematical morphology, PointSSIM enables robust comparison across binary images of varying resolutions by transforming them into marked point pattern representations. The key features of the image, referred to as anchor points, are extracted from binary images by identifying locally adaptive maxima from the minimal distance transform. Image comparisons are then performed using a summary vector, capturing intensity, connectivity, complexity, and structural attributes. Results show that this approach provides an efficient and reliable method for image comparison, particularly suited to applications requiring structural analysis across different resolutions.

## I. INTRODUCTION

Image comparison is a fundamental task in fields such as computer vision (Lowe, 2004; Szelisk, 2020), medical imaging (Litjens et al., 2017), and geostatistics (Pyrcz & Deutsch, 2014). Accurate and efficient image comparison metrics are essential for applications like image registration, quality assessment, and structural analysis (Wang & Bovik, 2009).

The most straightforward way of comparing images is pixel-to-pixel correspondence, such as Mean Squared Error (MSE) (Maindonald, 2007),

$$MSE(x,y) = \frac{1}{n_{rc}} \sum_{i=1}^{n_{rc}} (x_i - y_i)^2$$

In this case, $x$ and $y$ are images with $n_{rc}$ number of pixels. This is often used as a loss function for training Machine Learning (ML)-models and can be effective at image reconstruction for example in generative ML-models such as Variational Autoencoders (VAEs) (Kingma & Welling, 2019). However, for image-comparison it has limitations. It is particularly sensitive to rotations, and the comparison value can be difficult to interpret (Goodfellow, 2016).

A more robust approach to image comparison is to use summary statistics that are rotation invariant. Univariate statistics summarize and compare pixel frequencies. The most common univariate statistics are the mean and the variance:

$$\mu_x = \frac{1}{n_{rc}} \sum_{i=1}^{n_{rc}} x_i, \quad \sigma_x^2 = \frac{1}{n_{rc}-1} \sum_{i=1}^{n_{rc}} (x_i - \mu_x)^2.$$

Here, $\mu_x$ is the mean and $\sigma_x^2$ is the variance of the values in an image. Comparing mean and variance of images can be effective, although it does not take into consideration interactions between pixels, which can be critical in capturing structure in the image.

Second-order statistics are commonly used to capture interactions between pixels. In geostatistics, variograms are used to visualize and study second-order interactions (Pyrcz & Deutsch, 2014). The variogram is directly connected to the covariance and it quantifies spatial correlations at different lags. The formula for the semi-variogram (variogram divided by 2) for an image $x$ at lag distance $h$ is:

$$\gamma(h) = \frac{1}{2|N(h)|} \sum_{(i,j) \in N(h)} |x_i - x_j|^2,$$

where $N(h)$ is the collection of pixel pairs where the lag distance is $h$. Variograms of images can be compared at different lags, or one can take the MSE between two variograms at all lag distances. The variogram is limited to characterizing second-order interactions, but higher-order interaction terms may be relevant.

There is an array of metrics to characterize higher-order interactions (Grammer et al., 2020; Leuangthong et al., 2004; Lyster et al., 2004; Zuo et al., 2023). Many of them rely on scanning the image templates. For a single inspection at one location a limited number of immediate neighbors is used. In 2D the template typically comprises of 5 to 9 cells (template center plus immediate neighbors). By scanning the entire image with the template, we can calculate frequencies of specific template events and then compare these frequencies between images. Such methods are called n-point histograms (Boisvert et al., 2010; Deutsch & Pyrcz, 2013; Grammer et al., 2020; Pyrcz, 2016; Tahmasebi, 2018). The histogram part comes from the fact that we often put the specific template events and their associated frequencies in histograms. A common metric of n-point histograms for image-comparisons is the MSE over the template value frequencies. These n-point histograms can be effective, but they pose computational challenges if the number of pixels in the image or the template size is large, for example, if more than the adjacent neighbors are considered. Furthermore, as the template size increases the number of possible template values increases exponentially, making it difficult to summarize in histograms. This is a challenge, as we are often interested in larger scale structures than the 4-point or 8-point template.

To address this challenge a lot of methods have been proposed. Tan et al., (2014) proposed using multi-dimensional scaling (MDS) to cluster patterns based on a training image. For each cluster, the centroid was extracted, such that for each pattern in an image the closest centroid decides which cluster the pattern belongs to. In this way one can compare histograms of pattern clusters rather than every possible pattern for large templates. Honarkhah & Caers, (2010) proposed an adaptive template selection method based on elbow point detection on the entropy. They defined the entropy to be the information needed to encode a pattern. They additionally used MDS to cluster patterns. Zuo et al., (2023) proposed a pattern classification distribution method (PCDM) inspired by Honarkhah & Caers, (2010) and correlation-driven direct sampling (Zuo et al., 2019) to make adaptive templates. Pattern clusters were found using hierarchical clustering (Vichi et al., 2022). All these methods rely on dimensionality reduction techniques (Nanga et al., 2021). This can be effective at binning templates values into clusters or groups, making it easier to summarize in histograms. However, it becomes difficult to interpret (Tahmasebi, 2018). Lilleborge et. al., (2024) proposed a method based on counting 3D template patterns and looking at the probability of the counts being samples from the same distribution instead of relying on dimensional reduction techniques.

Other approaches for image comparison that has gained a lot of popularity recently are composite metrics. In particular, the Structural Similarity Index Measure (SSIM) proposed by Wang et al. (2004) provides a metric between 0 and 1 for structural similarity. Originally, it was constructed to measure image degradation as perceived changes in structural information, but it was later adopted as an image comparison metric. It works by assessing and comparing three measures of a pair of images. The measures are luminance; represented by the mean of the pixel values, contrast; represented by the variance of the pixel values and structure; represented by the covariance between the image pixels. Combining it all together the metric for two images $x$ and $y$ becomes:

$$SSIM(x,y) = \frac{(2\mu_x\mu_y + c_1)(2\sigma_{xy} + c_2)}{(\mu_x^2 + \mu_y^2 + c_1)(\sigma_x^2 + \sigma_y^2 + c_2)},$$

$\sigma_{xy} = \frac{\sum_{i=1}^{n_{rc}}(x_i-\mu_x)(y_i-\mu_y)}{n_{rc}-1}$ is the covariance between pixel values of images $x$ and $y$. Further, $c_1 = (0.01 \cdot L)^2$ and $c_2 = (0.03 \cdot L)^2$ stabilize the division with the weak denominator, with $L$ being the dynamic range of the pixel-values (Wang et al., 2004). The SSIM metric has notable weaknesses (Brunet et al., 2012). First, the metric is not resolution invariant, meaning that $x$ and $y$ must have the same pixel dimensions. Further, the structure measure in SSIM is particularly sensitive to pixel-level information, making it less robust as it computes the covariance at a pixel-to-pixel basis. Improvements such as Complex Wavelet SSIM (CW-SSIM), proposed by Sampat et al. (2009), which first applies the complex wavelet transform and then calculates SSIM on the transformed signals, yield better performance and higher scores for transformed images (Sampat et al., 2009). While effective for continuous images, wavelet transforms are unsuitable for discrete images like binary or segmented ones, as they assume smooth intensity variations. Another popular extension is the Multi-Scale SSIM (MS-SSIM) (Wang et al., 2003) that gives a more robust metric for comparing structures at different scales.

This paper introduces a novel image-to-image comparison metric that addresses these limitations, offering robustness to resolution and rotation. While theoretically invariant to resolution, some smoothing effects at higher resolutions introduce slight sensitivity to scale. Our approach, inspired by SSIM (Wang et al., 2004) and the principles of Mathematical Morphology (MM) developed by Matheron & Serra (2000), applies transformations to convert the image into a point process represented by anchor points, like how MM uses anchors to define points invariant under transformation. This allows us to bypass pixel-to-pixel correspondence. It measures and compares diverse aspects of the images, facilitating image comparison at different resolutions. Our proposed image-to-image comparison metric first transforms both images into a lower dimensional marked point-process representation (Ripley, 2014) where each point in addition to its location has two marks: radius and object label. We then do the comparison based only on this representation. The locations are found by extracting significant landmarks, called anchor points. These anchor points are determined through a novel locally adaptive maxima from the minimal distance transformation MM operator. This transform is resolution invariant, meaning it is not sensitive to pixel dimensions. Dimensionality reduction is accomplished by compressing the image with $n_{rc}$ dimensionality into a set of anchor points with $n_p \times 2$ dimensionality (2 coordinates) where $n_p \ll n_{rc}$ enabling efficient and robust comparison. The added marks of radius and label (described in Section 2) make the procedure output a marked point-process with dimensionality $n_p \times 4$. Once the point process is described, images are compared by evaluating four key measures related to the marked point-process,

- Anchor count: Number of anchor points
- Area coverage: a measure of the overall area spanned by the anchor point radii relative to the image size.
- Anchor points per object: a measure of the average heterogeneity of objects.
- Spatial variance irregularity: a measure of the spread of anchor points.

Notably, the measure comparison is rotation invariant.

Section 2 describes the methodology. Section 3 demonstrates the performance on multiple datasets and benchmarks our proposed PointSSIM metric against MSE, MS-SSIM, and SSIM image-to-image metrics. Section 4 reviews the results and their implications. We conclude with a summary of our findings and suggestions for future research directions in Section 5.

## II. METHODOLOGY

We introduce the methodology by first showing a schematic representation of the PointSSIM method, explaining each step of the process. We then develop the point-process representation using an example binary image. Figure 1 shows a schematic representation of the PointSSIM method. There are two steps: grid- to marked point-process representation and marked point-process representation to PointSSIM scalar value.

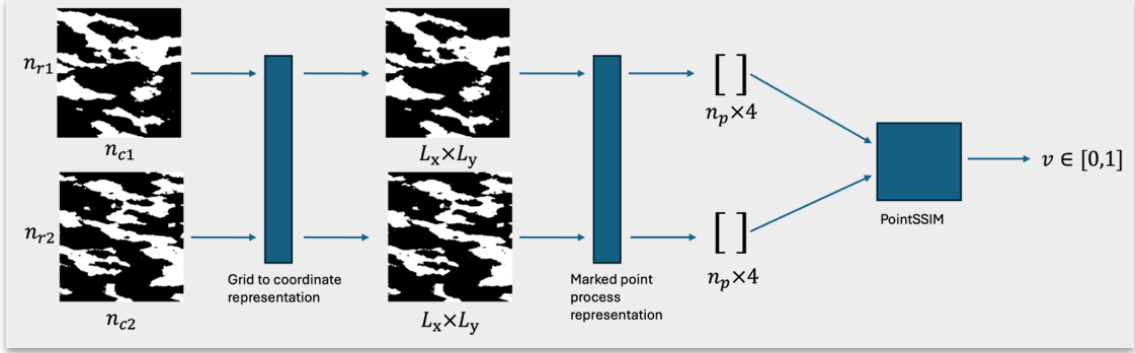

*Figure 1: Schematic representation of the PointSSIM method.*

### A. Grid- to marked-point process representation

The first step of the method is transforming the images from grid coordinates to a base coordinate system. If the two binary images have the same size, i.e. $n_{c1} = n_{c2}$ and $n_{r1} = n_{r2}$, then we can proceed to the next step. If not, we set $(L_x, L_y) = \min((n_{c1}, n_{r1}), (n_{c2}, n_{r2}))$ and $c_{xi} = n_{ci}\Delta x_i, c_{yi} = n_{ri}\Delta y_i, i \in [1,2]$, where $(c_{xi}, c_{yi})$ are the coordinates in the base coordinate system and $\Delta x_i = \frac{L_x}{n_{ci}}$, $\Delta y_i = \frac{L_y}{n_{ci}}$ are the cell sizes of the base coordinate system.

Once a base coordinate system is established, we can extract the anchor points of the images. We introduce this step of the PointSSIM method by showing the procedure to a marked point-process on a single example binary image $x = \{x_{ij}, i = 1,2, \ldots, \frac{L_x}{\Delta x}, j = 1,2, \ldots, \frac{L_y}{\Delta y}\}$. Figure 2 shows a display of the binary image produced by multiple point-statistics-based simulation (Grammer et al., 2020).

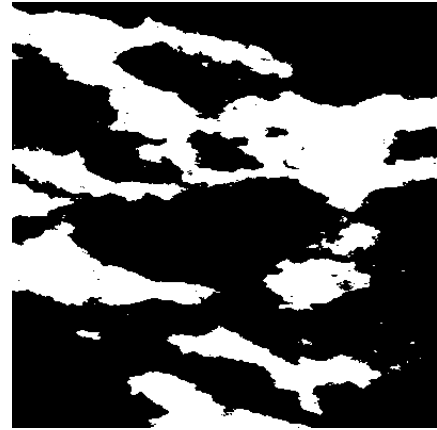

*Figure 2: Display of a 300x300 binary image produced using Multiple Point-Statistics.*

#### 1) Minimal Distance Transform

The first step involves performing a minimal distance transform on the image (Banerji, 2000). This transform is defined as follows:

*Equation 1*

$$D(x)_{ij} = \{\min_{(k,l)} d((i,j),(k,l)) : x_{kl} = 0\},$$

where $d((i,j),(k,l))$ is the Euclidean distance between two grid cell positions $(i,j)$ and $(k,l)$. For a grid cell where $x_{ij} = 0$ this distance is 0. For a grid cell where $x_{ij} = 1$, $D(x)_{ij}$ in Equation 1 is larger than 0. This transform calculates the minimal distance from any pixel to the nearest pixel with a value of 0. In Figure 3 we display the minimal distance transform of Figure 2.

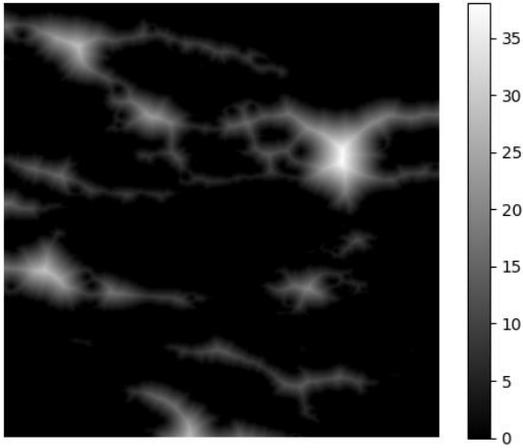

Figure 3: The minimal distance transform of the binary image in Figure 2.

2) Identifying Anchor points

A. The image $D(x)_{ij}, i = 1, \ldots, \frac{L_x}{\Delta x}, j = 1, \ldots, \frac{L_y}{\Delta y}$ has maximum values along the skeleton of the objects, which lie along ridges. The concept of anchor points involves finding these maximum values of the minimal distance transform, which can be viewed as points invariant under differentiation, analogous to anchor points in MM (Van Droogenbroeck, 2009).

B. For discrete images, local maxima can be identified using a template. Common choices include the 4-point and 8-point templates, depending on whether diagonals are considered neighbors. We apply the 8-point template, treating diagonals as neighbors. We also allow ties, meaning if any point is greater than or equal to all its 8-point neighbors, it is considered a local maximum. For an image $y$, the local maximum is defined as:

*Equation 2*

$$L(y)_{ij} = \begin{cases} 1 & \text{if } y_{ij} \geq y_{kl} \text{ for all } y_{kl} \in \mathcal{N}_{ij}, \\ 0 & \text{otherwise,} \end{cases}$$

C. where $\mathcal{N}_{ij} = \{i + a, j + b \mid a, b \in \{-1, 0, 1\} \wedge (a, b) \neq (0, 0)\}$ are the 8 neighbor grid cells of $(i, j)$.

Figure 4 highlights the local maxima within the distance transform $L(D(x))$ as red points.

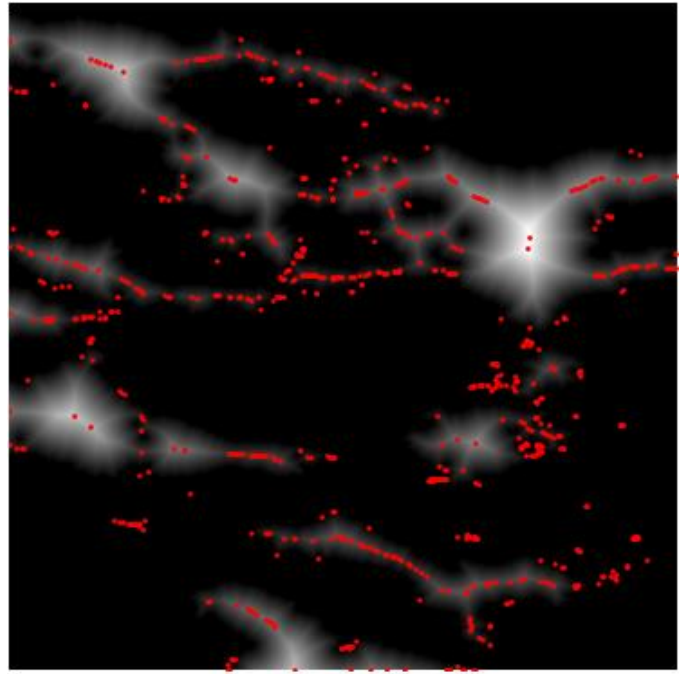

Figure 4: Local maximum points highlighted by red markers of the minimal distance transform of the binary image in Figure 2.

3) Locally Adaptive Anchor points

To avoid superfluous high-density points and to integrate local scale information, we use locally adaptive anchor points, ensuring that no two anchor points are closer to each other than to the edge of the object:

*Equation 3*

$$L'(y)_{ij} = \begin{cases} 1 & \text{if } D(y)_{ij} \leq \min_{(k,l)} \{d((i,j),(k,l)) : L(y)_{kl} = 1\} \\ 0 & \text{otherwise} \end{cases}.$$

Figure 5 shows the customized anchor points $L'(D(x))$ with both the distance transformed image and the original image as background.

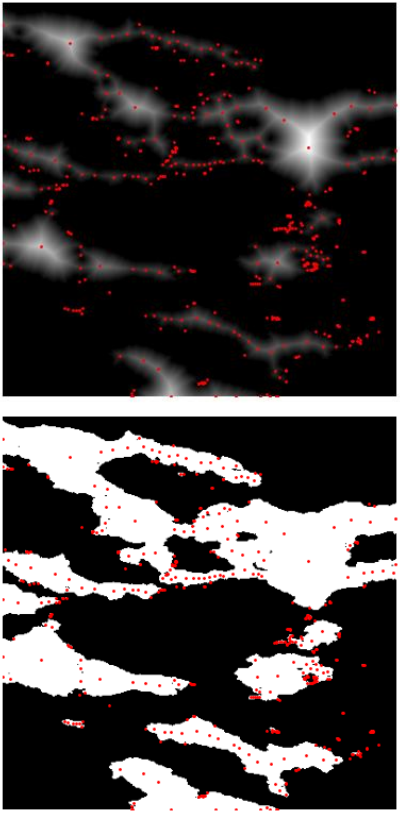

*Figure 5: Locally adaptive local maximum points of the minimal distance transform of the binary image in Figure 2 highlighted with red markers, with the minimal distance transform as background to the left and the original image as background to the right.*

The locally adaptive method described in Equation 3 ensures that anchor points are distributed based on object size, preserving the relative location and number of points for objects of the same shape, regardless of scale. This approach maintains sufficient spacing between anchor points while preserving enough density to capture meaningful structural details.

B. *Marked Point-Process Representation*

Once the anchor points are identified, the image is effectively converted into a low-dimensional point-process representation. The point-process representation has a designed feature in that no two anchor points can be closer to each other than to the border of the object. This can be viewed as a marked point-process, where each anchor point has two marks. The first is the effective radius of the anchor point. If we draw a circle around each anchor point to the border of the object, we have the property that each circle only contains a single point, the center. The radius is an important mark as it tells us about the local regularity around the anchor point. Figure 7 shows an illustration of the binary image with the anchor point circles included. The second mark is the object label of the anchor point, telling us which object the anchor point belongs to. Object labels are found by a standard connected component method(Virtanen et al., 2020). This method is using a structuring element to scan the image and labeling based on the labels of its neighbors. We use the 8-neighbors as the structuring element. This is highlighted by the color in Figure 7.

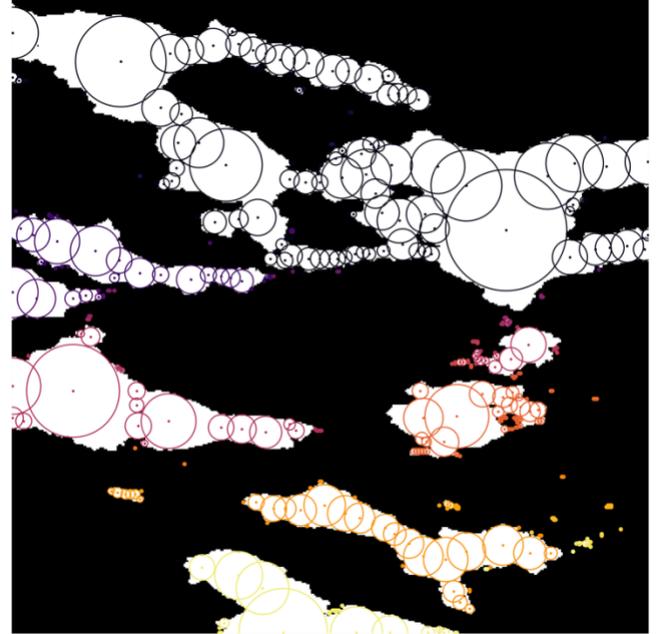

*Figure 6: Anchor points of the binary image presented in Figure 2, with the inclusion of position (represented as dots), circles (with the anchor point positions as origo, and the radius representing the distance to the closest border of the object), and label (represented by color scheme).*

Drawing inspiration from the SSIM we focus on properties of the marked point-process that are invariant under rotation and resolution scaling transformations. We introduce the following anchor point notation:

*Equation 4*
$$A^p = \{(i,j): L'(D(x))_{ij} = 1\},$$

*Equation 5*
$$A^r = \{D(x)_{ij}: L'(D(x))_{ij} = 1\},$$

*Equation 6*
$$A^l = \{l_1, l_2, \ldots, l_{|A^p|}\},$$

where $A^p$ represents the grid coordinates of the marked point-process, $A^r$ represents the corresponding radii of the marked point-process, while $A^l$ is a collection of the anchor point labels corresponding to the unique connected areas in the image (color coded in Figure 7). Putting the coordinates and the 2 marks together, we end up with an $n_p \times 4$ vector representation of the binary image: $[A_i^p, A_j^p, A^r, A^l]$.

## C. Marked point-process representation to PointSSIM scalar

Given the marked point-process vector provided in the previous section we present four measures that capture features of the binary image, which help construct the PointSSIM metric.

**Anchor count**

The first measure is calculated as the number of unique anchor points in the image,

*Equation 7*

$$V_1(x) = |A^p|.$$

It summarizes the intensity of points in the image.

**Area coverage**

The second measure counts the sum of the squared radius marks (proportional to the area of the circle) for the marked point-process, and scales against the area of the image, giving a dimensionless measure. This global measure relates to the foreground proportion in the image (+ some circle overlap),

*Equation 8*

$$V_2(x) = \frac{\sum_{i=1}^{|A^r|} A_i^{r\,2}}{L_x \cdot L_y}.$$

**Anchor points per object**

The third measure evaluates the average number of anchor points per object:

*Equation 9*

$$V_3(x) = \frac{V_1(x)}{\max(A^l)}.$$

Here, the number of objects in the image is $\max(A^l)$, as the objects are ordered from 1 to the number of objects. This measure is related to the heterogeneity of objects in the image. When this number is low it indicates homogeneous objects, a high number indicates heterogeneous objects.

**Spatial variance irregularity**

The first three measures capture the basic parameterization of the marked point-process. The fourth measure assesses the point structure, focusing on the spatial correlation of the anchor points. This measure provides insights into the clustering, randomness, and structure of the points. Given how the marked point-process is defined in this context, classical measures of spatial point clustering, such as Moran's I (Moran, 1950), are inappropriate. This is because Moran's I is designed for assessing spatial clustering of points, whereas in this context, anchor points are inherently separated due to the repulsive effects of local adaptivity and the separation of objects. Therefore, we need to define clustering differently.

We compare the distribution of points to a Poisson point-process (Daley & Vere-Jones, 1990). In a Poisson point-process, the number of points in an area $B$ follows a Poisson distribution with mean $\lambda|B|$, where $\lambda$ is the intensity of points. It can be assessed by $\lambda = \frac{n}{|A|}$, where $n$ is the total number of points in the domain, and $|A|$ is the area of the domain.

Given the theoretical variance, we compare it to the empirical variance by splitting the entire domain into several disjoint quadratic subregions B and then counting the number of anchor points within these. In this work, we use 100 subregions, meaning we split our domain into 10x10 subregions, which seems to work well empirically.

We compute the empirical variance for all such counts. To normalize the measure, we take the difference between the empirical variance and the theoretical variance, divided by their sum. This gives a measure that can take values in the range $[-1,1]$. To normalize the measure to the range $[0,1]$ we add 1 and divide by 2. This simplifies to the following measure:

*Equation 10*

$$V_4(x) = \frac{1}{1 + \frac{\lambda|B|}{s^2}},$$

where:
- $s^2$ is the empirical variance in the number of anchor points within subregions,
- $\lambda|B| = \frac{n}{100}$ is the theoretical variance,
- $|B|$ is the area of a quadratic subregion, which is $\frac{L_x L_y}{100}$.

We calculate $s^2$ as,

*Equation 10*

$$s^2 = \frac{1}{n}\sum_{i=1}^{n}(n_i - \bar{n}_i)^2,$$

here $n_i$ is the collection of anchor points in subregion $i$.

Figure 8 is an illustration of the measure for 3 different binary images. When the 1-valued pixels are evenly spaced, the variance between subregion counts is 0 and the measure is 0, if the 1-valued pixels are randomly placed the measured and expected variance is the same, giving a value close to 0.5, and if the 1-valued pixels are clustered the measured variance is much larger than the expected such that the contribution from the expected variance becomes negligible and the measure gets close to 1.

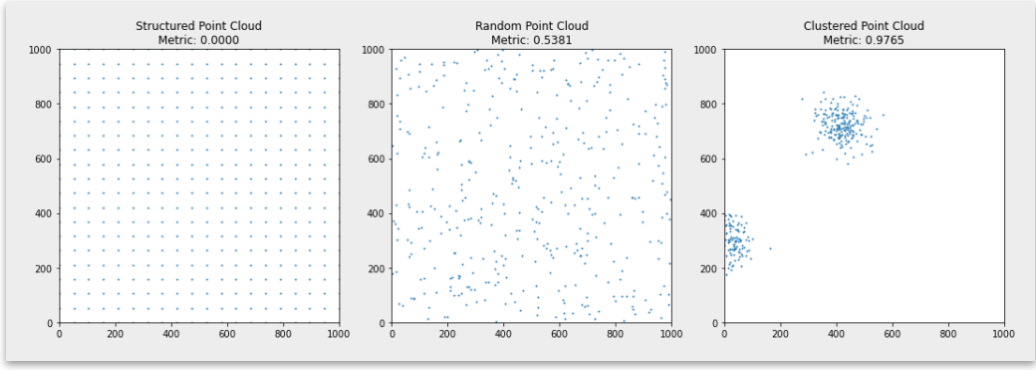

*Figure 7: Illustration of the variance irregularity measure for three different binary images. The left display represents a structured arrangement of points, the middle image represents a random arrangement of points and the image to the right represents a clustered arrangement of points. Each of the images have a structure score (ranging from 0 to 1) representing the degree of clustering in the image.*

Given the four measures, we represent the binary images as 4-dimensional vectors that can be compared among each other to capture structural similarity with the PointSSIM metric. We employ a reference-free form of comparison, like in SSIM (Wang et al. 2004). The Euclidean distance between each measure is compared and normalized by the maximum value to yield a value between zero and one. The normalized distance of each measure is then averaged to provide a comparison value. For comparing two images $x_1$ and $x_2$ we have metric:

*Equation 11*

$$PointSSIM(x_1, x_2) = 1 - \frac{1}{4}\left(\sum_{i=1}^{3} \frac{(V_i(x_1) - V_i(x_2))^2}{\max(V_i(x_1), V_i(x_2))^2} + (V_4(x_1) - V_4(x_2))^2\right),$$

we avoid double normalization of $V_4(\cdot)$ by putting it outside the sum. This metric gives values in the range [0,1], where 0 represents no structural similarity, and 1 represents full structural similarity in the marked point-process vector space.

### III. RESULTS

*A. Test Image Scenarios*

To assess the performance of the proposed PointSSIM metric, we evaluated it across a range of simulated binary images designed to capture various structural patterns. The primary goal was to see how well PointSSIM captures the similarities and differences between images of different types and structures, particularly in scenarios where traditional methods might struggle. One key consideration is the number of anchor points: PointSSIM relies on enough anchor points being present in the image to generate meaningful statistical comparisons. For example, if large continuous objects cover most of the image, there may be too few anchor points to reliably capture structural differences, leading to increased variance in the metric.

We considered five distinct types of image scenarios, each containing multiple objects. We generated 50 realizations for each dataset to evaluate the performance of PointSSIM. The datasets used for comparison include:

- MPS images: These were generated using Multiple-Point Statistics to emulate geological binary surfaces.

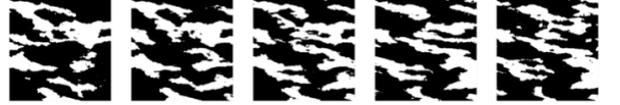

*Figure 8: Multiple-Point Statistics binary images.*

- TGRF images: Truncated Gaussian Random Field realizations were used, which include nested variograms.

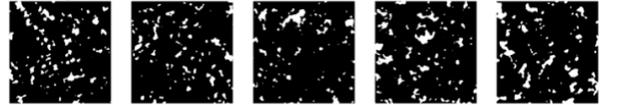

*Figure 9: Truncated Gaussian Random Field images.*

- Structured ellipses images: These images consist of ellipses arranged in a grid in a highly structured manner.

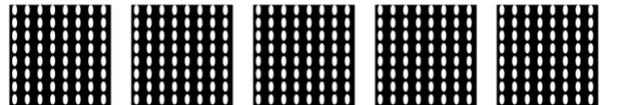

*Figure 10: Structured ellipses images.*

- Distorted ellipses images: These images feature ellipses placed randomly in a grid with some

added Gaussian noise to distort their shapes.

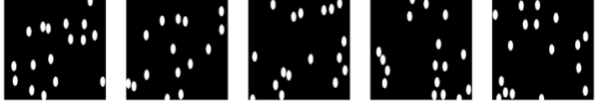

*Figure 11: Randomly placed ellipses images.*

- Mixture of ellipses and circles: These images include a mix of ellipses and circles with overlapping objects, restricted to certain areas of the image (corners).

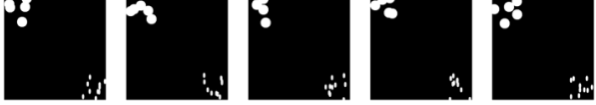

*Figure 12: Mixture of circles and ellipses images, allowing for overlap and restricted to corners of the image.*

These scenarios were chosen to test how well PointSSIM differentiates between images with varying structural complexities, object arrangements, and noise levels. The ability to handle both structured and random configurations is critical for geostatistical applications, where image structures can vary significantly.

### B. Results with Test Images

Figure 14 shows histograms and scatterplots of all four key measures for each image scenario, allowing us to observe the separation between datasets. From these plots, PointSSIM effectively distinguishes between the different image scenarios, and the measures themselves are not redundant, meaning no two measures have strong correlations. This independence between measures is important because it ensures that PointSSIM captures a variety of structural aspects of the images, rather than over-emphasizing a single characteristic.

In particular, the structured ellipses dataset shows zero variance in the measures across its 50 realizations. This is as expected because each realization is a direct copy of the original structure. By contrast, both the TGRF and MPS datasets display much higher variance, reflecting the greater diversity in structures between individual realizations. This result highlights how PointSSIM can effectively capture and quantify both structured and irregular spatial patterns.

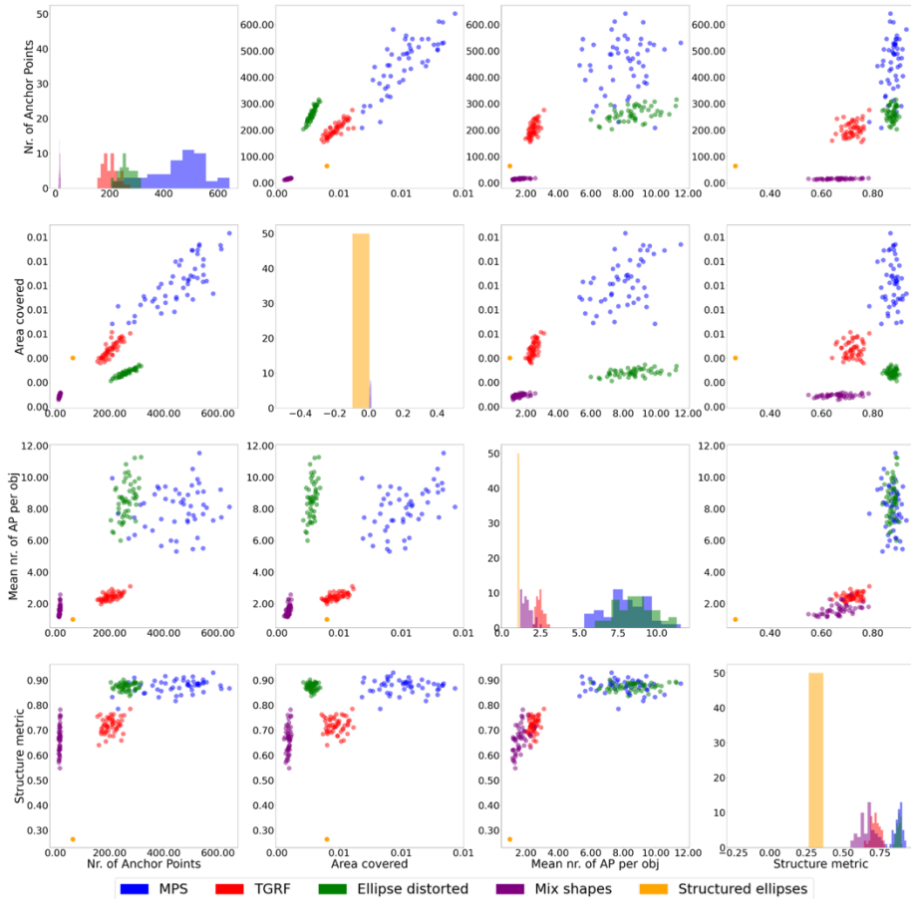

*Figure 13: Histograms of all 5 datasets (Figure 9-13) for each of the measures along the diagonals of the Figure. The rows represent the combinations of two measures with the individual data points represented as colored markers.*

In Figure 15, we compare PointSSIM against three other popular metrics: SSIM, MSE, and MS-SSIM. The violin plots show the distribution of metric values for pairwise comparisons between the different datasets. Each subplot represents a different dataset combination, with PointSSIM, SSIM, MSE, and Multiscale-SSIM values plotted in blue, orange, green, and red, respectively.

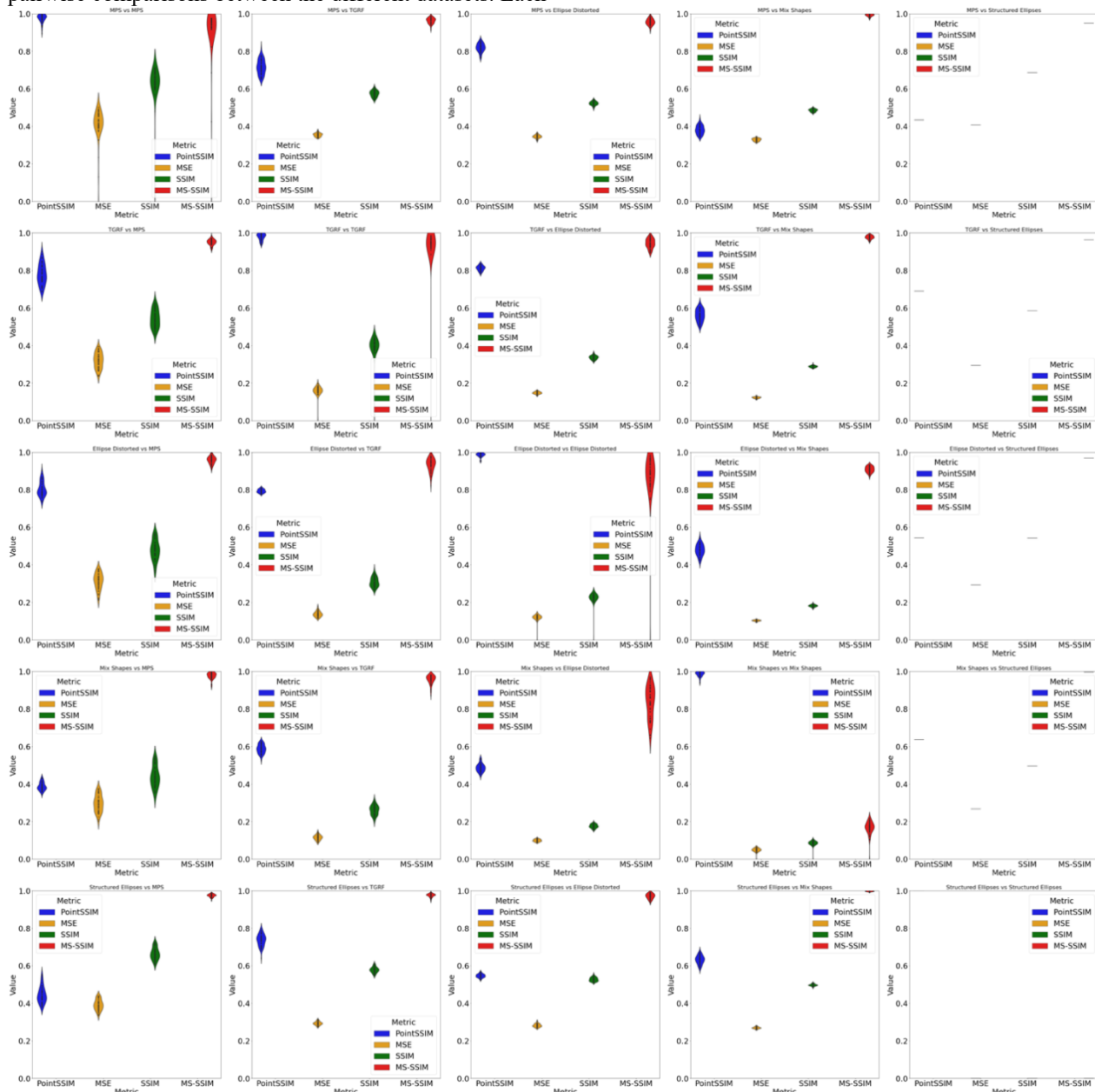

Figure 15: The violin plot of the distribution of the PointSSIM (blue), MSE (yellow), SSIM (green) and MS-SSIM (red) values for each combination of datasets (Figure 9-13). Each combination of datasets are individual subplots, with a unique color for each metric.

Several important observations can be made from this figure:

- Within-class similarity: For each dataset, PointSSIM consistently returns a value close to 1 for within-class comparisons (diagonal elements in the display), with minimal variance. This high within-class similarity indicates that PointSSIM can recognize and quantify structural consistency within these datasets more effectively than the other metrics. By contrast, SSIM, MSE, and MS-SSIM often show lower within-class similarity, with greater variance, making them less precise.

- Between-class differentiation: PointSSIM also excels at distinguishing between different image classes. In many cases, PointSSIM produces lower between-class similarity scores than SSIM, MSE, or MS-SSIM. This superior ability to differentiate between distinct datasets is critical for applications where accurate discrimination between image types is required, such as in geological modeling or pattern recognition.

- Responsiveness to dataset combinations: In the first column of the violin plots, PointSSIM demonstrates a clear response to different combinations of datasets, with varying means and variances. This contrasts with the other metrics, which tend to produce similar responses for different dataset combinations, making it harder to distinguish between them. This indicates that PointSSIM is more sensitive to structural differences, allowing for finer-grained comparisons.

C. *Additional Resolution Experiment*

One of the main advantages of PointSSIM is its resolution invariance, meaning the metric remains effective even when images are rescaled. This property is especially useful in geostatistics, where images of different resolutions need to be compared. To test this, we conducted an experiment where the mixture of ellipses and circles dataset was generated at three different resolutions: 256x256, 512x512, and 1024x1024. For each resolution, 50 realizations were generated, and the results were compared.

Figure 16 shows five realizations for each resolution, illustrating how the objects are scaled across different resolutions. As the resolution increases, the edges of the objects become smoother, which naturally reduces the number of local maxima detected in the minimal distance transform. This reduction in anchor points could potentially affect the metric, but the PointSSIM method adapts well to these changes.

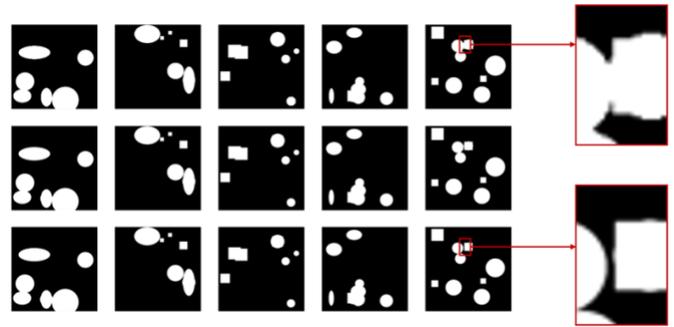

Figure 16: 5 realizations of 256x256, 512x512 and 1024x1024 resolution images.

The histograms in Figure 17 display the measures for the three resolutions, confirming that the values overlap significantly across different resolutions. Figure 18 shows the individual images as scatter points for all measures, with the low and high resolution on the $x$- and $y$- axis. Ideally, all the scatter points would lie on the line $y = x$. We observe that there are some fluctuations from this line, especially for the 3$^{rd}$ and 4$^{th}$ measure, where some points lie below the diagonal line. This is an effect of smoothing, where low resolution images have fewer nr. of pixels that can cause objects to merge as in Figure 17. Since measure 3 is the same as measure 1 except that we divide by the number of objects, this causes the low-resolution images to have a lower value when objects are merged. While some small fluctuations are observed due to the smoothing effect between realizations, the measures consistently capture the structural integrity of the images, regardless of resolution. This demonstrates the robustness of the method in scenarios where pixel resolution varies, a significant improvement over pixel-based methods like MSE and SSIM, which are resolution-dependent.

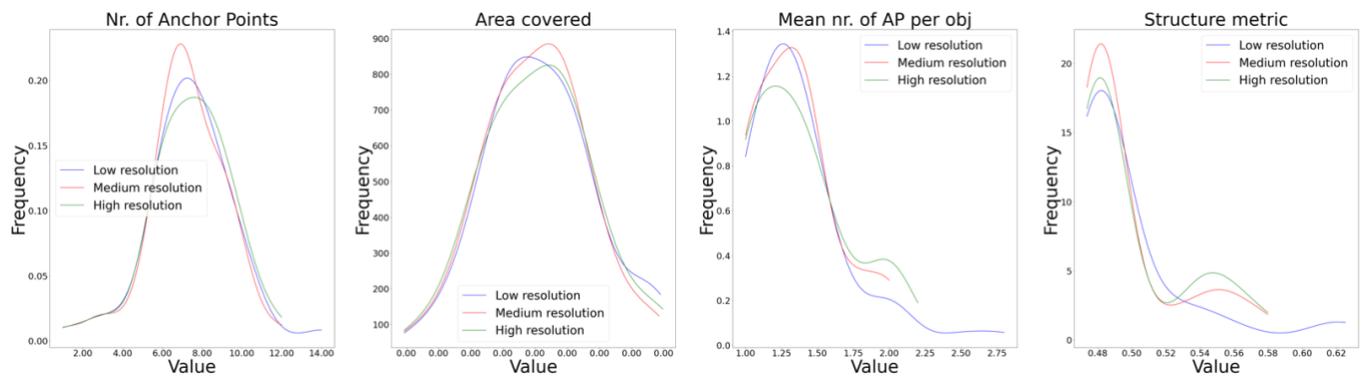

Figure 17: Histograms of all 3 datasets of different resolution (Figure 17) for each of the measures.

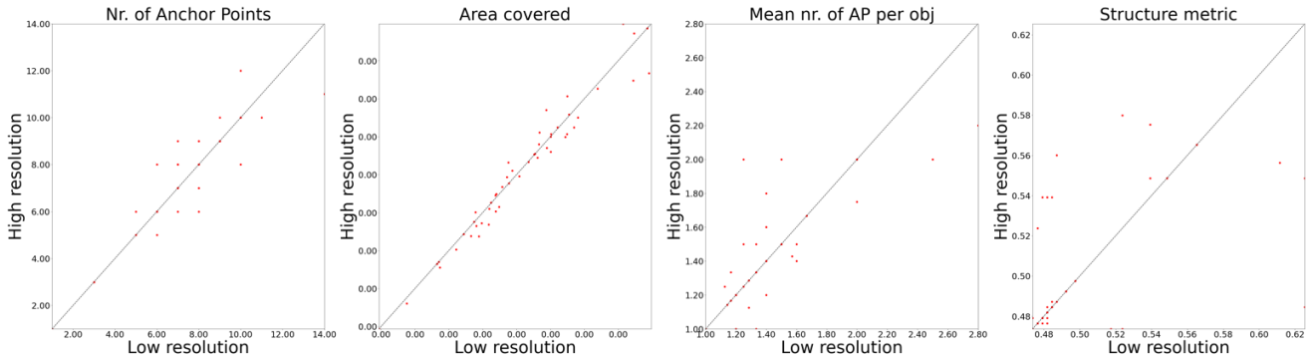
*Figure 18: Scatter plot of each measure for low vs high resolution.*

## IV. DISCUSSION

The results demonstrate that PointSSIM effectively distinguishes image scenarios, outperforming common metrics like SSIM, MSE, and MS-SSIM. By comparing point processes, we can compare across resolutions and create a rotation invariant measure.

Compressing a binary image into four summary measures inevitably leads to some loss of detail. For example, two datasets with different object shapes (e.g., curved vs. straight in Figure 19 and 20) may not be fully distinguished by the current metric, as the curvature is not explicitly captured. This highlights the somewhat arbitrary nature of the chosen measures. While the four measures (anchor count, area coverage, anchor points per object, and spatial variance irregularity) provide a good summary for these datasets, other measures could be more appropriate in different contexts, depending on the specific structural features of interest.

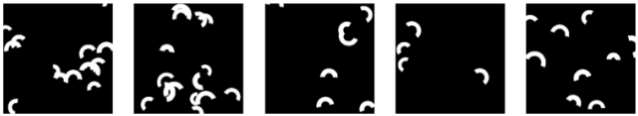
*Figure 19: Binary images of curved objects*

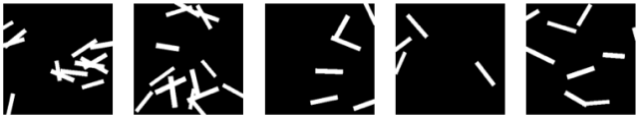
*Figure 20: Binary images of straight objects*

Additionally, relying on anchor points requires a sufficient number of these points to generate meaningful statistics. When there are too few anchor points, especially in images with large, continuous objects, the point-based summaries become overly sensitive and less reliable. This can lead to reduced robustness in distinguishing between images with subtle structural differences.

The transformation of binary images into anchor points is efficient, but it involves a trade-off between detail and computational speed. While PointSSIM is effective for binary images, alternative methods like CW-SSIM (Sampat et al., 2009), which maps images to the frequency domain, may be better suited for more complex image types like RGB images, where finer pixel-level details matter.

Despite these limitations, the flexibility of PointSSIM is a strength. The framework allows for adjustments at each stage. The measures derived from the point process can be tailored to look for specific features relevant for the task at hand. Likewise, the extraction of points and marks can be refined, and the choice of marks could also be different from what we have used here. This makes the core idea very adaptable for a wide range of both datasets and applications.

## V. CONCLUSION

In this work, we introduced PointSSIM, a novel, low-dimensional image-to-image comparison metric designed to be invariant to resolution and rotation. The metric compresses complex binary images into a marked point-process representation, allowing us to capture essential structural information efficiently. By utilizing four key measures—anchor point intensity, area coverage, anchor points per object, and anchor autocorrelation—PointSSIM offers a robust and scalable method for comparing binary images across a range of structural scenarios. Our evaluations show that PointSSIM not only outperforms popular comparison metrics like SSIM, MSE, and MS-SSIM but also exhibits the critical advantage of being resolution invariant.

Despite its strengths, PointSSIM has limitations. The compression of binary images into four summary measures inherently leads to some loss of information, meaning that certain image characteristics, such as fine geometric details or subtle shape variations (e.g., curvature differences), may not be fully captured. The method works best when there are enough anchor points in the image, which typically requires a reasonable number of objects or distinct features. In images where the number of anchor points is low, or when large continuous objects dominate the scene, the resulting point-based summaries may become overly sensitive to minor variations, reducing the robustness of the comparisons.

While we have focused on four specific measures for PointSSIM, these are not exhaustive. The choice of these measures reflects a balance between efficiency and structural descriptiveness, but other measures could be explored to capture additional image properties, depending on the specific application. For example, integrating measures that better account for curvature, texture, or more complex spatial interactions could enhance the method's ability to distinguish between images with more intricate differences. However, incorporating additional measures must be done with caution to avoid redundancy and unnecessary complexity, as this could reduce interpretability and make the method harder to apply consistently.

A valuable extension of this work would be to experiment with alternative transformations and measures that go beyond binary images, applying the PointSSIM framework to grayscale or RGB images. In such cases, anchor points could be selected based on other criteria, such as intensity gradients or color homogeneity, and then compared using an expanded set of measures. This would open the method to a broader range of applications, including medical imaging, geological analysis, and remote sensing, where preserving structure across scales is crucial.

Additionally, the PointSSIM framework could serve as a foundational tool in machine learning contexts, particularly for training generative models. By incorporating structural measures alongside pixel-level accuracy, PointSSIM could act as a regularization term, ensuring that the structural integrity of generated images is preserved. This is particularly valuable in tasks such as image synthesis, where maintaining the underlying geometry or patterns of the training data is critical to the quality of the generated outputs.

In summary, PointSSIM offers an efficient, scalable, and flexible approach to image comparison, particularly for binary images requiring structural analysis. Its resolution invariance, combined with its ability to represent images in a low-dimensional space, makes it a valuable tool for a wide range of geostatistical and image analysis applications. Future research should explore more sophisticated transformations and measures to further expand its applicability and utility, particularly in handling more complex datasets.

REFERENCES

Banerji, A. (2000). An introduction to image analysis using mathematical morphology. In *IEEE Engineering in Medicine and Biology Magazine* (Vol. 19, Issue 4).

Boisvert, J. B., Pyrcz, M. J., & Deutsch, C. V. (2010). Multiple point metrics to assess categorical variable models. *Natural Resources Research*, *19*(3). https://doi.org/10.1007/s11053-010-9120-2

Brunet, D., Vrscay, E. R., & Wang, Z. (2012). On the mathematical properties of the structural similarity index. *IEEE Transactions on Image Processing*, *21*(4). https://doi.org/10.1109/TIP.2011.2173206

Deutsch, C. V, & Pyrcz, M. J. (2013). A Review and Teachers Aide on Multiple Point Statistics. *Center for Computational Geostatistics*, *15*.

Pyrcz and Deutsch (2014): Geostatistical Reservoir Modeling. Oxford University Press, USA.

Grammer, G. M., Harris, P. M. "Mitch," & Eberli, G. P. (2020). Multiple-point Geostatistics. In *Integration of Outcrop and Modern Analogs in Reservoir Modeling*. https://doi.org/10.1306/m80924c18

Honarkhah, M., & Caers, J. (2010). Stochastic simulation of patterns using distance-based pattern modeling. *Mathematical Geosciences*, *42*(5). https://doi.org/10.1007/s11004-010-9276-7

Kingma, D. P., & Welling, M. (2019). An introduction to variational autoencoders. In *Foundations and Trends in Machine Learning* (Vol. 12, Issue 4). https://doi.org/10.1561/2200000056

Learning, D. (2016). Deep Learning - Goodfellow. *Nature*, *26*(7553).

Leuangthong, O., McLennan, J. A., & Deutsch, C. V. (2004). Minimum acceptance criteria for geostatistical realizations. *Natural Resources Research*, *13*(3). https://doi.org/10.1023/B:NARR.0000046916.91703.bb

Litjens, G., Kooi, T., Bejnordi, B. E., Setio, A. A. A., Ciompi, F., Ghafoorian, M., van der Laak, J. A. W. M., van Ginneken, B., & Sánchez, C. I. (2017). A survey on deep learning in medical image analysis. In *Medical Image Analysis* (Vol. 42). https://doi.org/10.1016/j.media.2017.07.005

Lowe, D. G. (2004). Distinctive image features from scale-invariant keypoints. *International Journal of Computer Vision*, *60*(2). https://doi.org/10.1023/B:VISI.0000029664.99615.94

Lyster, S., Ortiz, J. C., & Deutsch, C. V. (2004). Scaling Multiple Point Statistics to Different Histograms. *Center for Computational Geostatistics Annual Report Papers*.

Maindonald, J. (2007). Pattern Recognition and Machine Learning. *Journal of Statistical Software*, *17*(Book Review 5). https://doi.org/10.18637/jss.v017.b05

Matheron, G., & Serra, J. (2000). The Birth of Mathematical Morphology. *Context*, *June*.

Moran, P. A. (1950). Notes on continuous stochastic phenomena. *Biometrika*, *37*(1–2). https://doi.org/10.1093/biomet/37.1-2.17

Nanga, S., Bawah, A. T., Acquaye, B. A., Billa, M.-I., Baeta, F. D., Odai, N. A., Obeng, S. K., & Nsiah, A. D. (2021). Review of Dimension Reduction Methods. *Journal of Data Analysis and Information Processing*, *09*(03). https://doi.org/10.4236/jdaip.2021.93013

Mariethoz and Caers (2014): Multiple-Point Geostatistics. *John Wiley & Sons*.

Ripley, B. D. (2014). Pattern recognition and neural networks. In *Pattern Recognition and Neural Networks*. https://doi.org/10.1017/CBO9780511812651

Sampat, M. P., Wang, Z., Gupta, S., Bovik, A. C., & Markey, M. K. (2009). Complex wavelet structural similarity: A new image similarity index. *IEEE Transactions on Image Processing*, *18*(11). https://doi.org/10.1109/TIP.2009.2025923

Szelisk, R. (2020). Computer Vision: Algorithms and Applications. In *Algorithms and applications* (Vol. 42).

Tahmasebi, P. (2018). Multiple point statistics: A review. In *Handbook of Mathematical Geosciences: Fifty Years of IAMG*. https://doi.org/10.1007/978-3-319-78999-6_30

Tan, X., Tahmasebi, P., & Caers, J. (2014). Comparing training-image based algorithms using an analysis of distance. *Mathematical Geosciences*, *46*(2). https://doi.org/10.1007/s11004-013-9482-1

Van Droogenbroeck, M. (2009). Anchors of Morphological Operators and Algebraic Openings. In *Advances in Imaging and Electron Physics* (Vol. 158). https://doi.org/10.1016/S1076-5670(09)00010-X

Vichi, M., Cavicchia, C., & Groenen, P. J. F. (2022). Hierarchical Means Clustering. *Journal of Classification*, *39*(3). https://doi.org/10.1007/s00357-022-09419-7

Virtanen, P., Gommers, R., Oliphant, T. E., Haberland, M., Reddy, T., Cournapeau, D., Burovski, E., Peterson, P., Weckesser, W., Bright, J., van der Walt, S. J., Brett, M., Wilson, J., Millman, K. J., Mayorov, N., Nelson, A. R. J., Jones, E., Kern, R., Larson, E., … Vázquez-Baeza, Y. (2020). SciPy 1.0: fundamental algorithms for scientific computing in Python. *Nature Methods*, *17*(3). https://doi.org/10.1038/s41592-019-0686-2

Wang, Z., & Bovik, A. C. (2009). Mean squared error: Lot it or leave it? A new look at signal fidelity measures. *IEEE Signal Processing Magazine*, *26*(1). https://doi.org/10.1109/MSP.2008.930649

Wang, Z., Bovik, A. C., Sheikh, H. R., & Simoncelli, E. P. (2004). Image quality assessment: From error visibility to structural similarity. *IEEE Transactions on Image Processing*, *13*(4). https://doi.org/10.1109/TIP.2003.819861

Wang, Z., Simoncelli, E. P., & Bovik, A. C. (2003). Multi-scale structural similarity for image quality assessment. *Conference Record of the Asilomar Conference on Signals, Systems and Computers*, *2*. https://doi.org/10.1109/acssc.2003.1292216

Zuo, C., Li, Z., Dai, Z., Wang, X., & Wang, Y. (2023). A Pattern Classification Distribution Method for Geostatistical Modeling Evaluation and Uncertainty Quantification. *Remote Sensing*, *15*(11). https://doi.org/10.3390/rs15112708

Zuo, C., Pan, Z., Gao, Z., & Gao, J. (2019). Correlation-driven direct sampling method for geostatistical simulation and training image evaluation. *Physical Review E*, *99*(5). https://doi.org/10.1103/PhysRevE.99.053310

Daley, D. J., & Vere-Jones, D. (1990). An Introduction to the Theory of Point Processes. *Journal of the American Statistical Association*, *85*(409). https://doi.org/10.2307/2289568

Lilleborge, M., Hauge, R., Fjellvoll,B. & Abrahamsen,P. (2024). Using Pattern Counts to Quantify the Difference Between a Pair of Three-Dimensional Realizations. *Mathematical Geosciences*. https://doi.org/10.1007/s11004-024-10145-6

AUTHORS

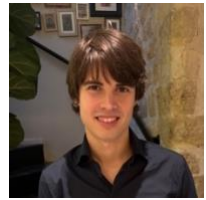

Oscar Ovanger
Norwegian University of Science and Technology
oscar.ovanger@ntnu.no
Alfred Getz' vei 1, 7034 Trondheim
Norway

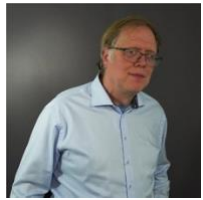

Dr. Ragnar Hauge
Norwegian Computing Center
hauge@nr.no
Gaustadalléen 23A, 0373 Oslo, Norway

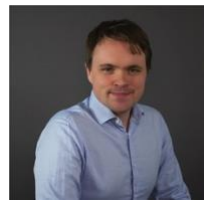

Dr. Jacob Skauvold
Norwegian Computing Center
jas@nr.no
Gaustadalléen 23A, 0373 Oslo, Norway

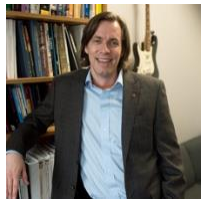

Prof. Michael Pyrcz
UT Austin
mpyrcz@austin.utexas.edu
Chemical and Petroleum Engineering, 200 E Dean Keeton St, Austin, TX 78712

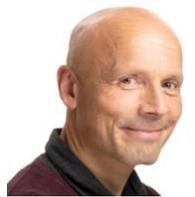

Prof. Jo Eidsvik
Norwegian University of Science and Techonology
jo.eidsvik@ntnu.no
Alfred Getz' vei 1, 7034 Trondheim
Norway